\documentclass[sigconf]{acmart}
\usepackage{multirow}
\usepackage{graphicx} 
\usepackage{subcaption} 
\usepackage{listings}
\usepackage{balance}
\usepackage{tcolorbox}
\tcbset{colback=gray!5, colframe=black!40, boxrule=0.4pt, arc=3pt}
\AtBeginDocument{%
  }

\copyrightyear{2026}
\acmYear{2026}
\setcopyright{cc}
\setcctype{by}
\acmConference[WWW '26]{Proceedings of the ACM Web Conference 2026}{April 13--17, 2026}{Dubai, United Arab Emirates}
\acmBooktitle{Proceedings of the ACM Web Conference 2026 (WWW '26), April 13--17, 2026, Dubai, United Arab Emirates}
\acmPrice{}
\acmDOI{10.1145/3774904.3792636}
\acmISBN{979-8-4007-2307-0/2026/04}

\begin{document}

\title{D-Models and E-Models: Diversity–Stability Trade-offs in the Sampling Behavior of Large Language Models}


 

\author{Jia Gu}
\email{gujia24s@ict.ac.cn}
\orcid{0009-0005-9532-4589}
\affiliation{%
  \institution{State Key Laboratory of AI Safety, Institute of Computing Technology, Chinese Academy of Sciences}
  \institution{University of Chinese Academy of Sciences}
  \city{Beijing}
  \country{China}
}

\author{Liang Pang*}
\email{pangliang@ict.ac.cn}
\orcid{0000-0003-1161-8546}
\affiliation{%
  \institution{State Key Laboratory of AI Safety, Institute of Computing Technology, Chinese Academy of Sciences}
  \institution{University of Chinese Academy of Sciences}
  \city{Beijing}
  \country{China}
}

\author{Huawei Shen}
\email{shenhuawei@ict.ac.cn}
\orcid{0000-0002-1081-8119}
\affiliation{%
  \institution{State Key Laboratory of AI Safety, Institute of Computing Technology, Chinese Academy of Sciences}
  \institution{University of Chinese Academy of Sciences}
  \city{Beijing}
  \country{China}
}

\author{Xueqi Cheng}
\email{cxq@ict.ac.cn}
\orcid{0000-0002-5201-8195}
\affiliation{%
  \institution{State Key Laboratory of AI Safety, Institute of Computing Technology, Chinese Academy of Sciences}
  \institution{University of Chinese Academy of Sciences}
  \city{Beijing}
  \country{China}
}

\thanks{*Corresponding author} 
\renewcommand{\shortauthors}{Gu et al.}

\begin{abstract}
The predictive probability of the next token (\(P_\text{token}\)) in large language models (LLMs) is inextricably linked to the probability of relevance for the next piece of information, the purchase probability of the next product, and the execution probability of the next action—all of which fall under the scope of the task-level target distribution (\(P_\text{task}\)). While LLMs are known to generate samples that approximate real-world distributions, whether their fine-grained sampling probabilities faithfully align with task requirements remains an open question. Through controlled distribution-sampling simulations, we uncover a striking dichotomy in LLM behavior, distinguishing two model types: D-models (e.g. Qwen-2.5), whose $P_\text{token}$ exhibits large step-to-step variability and poor alignment with $P_\text{task}$; and E-models (e.g. Mistral-Small), whose $P_\text{token}$ is more stable and better aligned with $P_\text{task}$. We further evaluate these two model types in downstream tasks such as code generation and recommendation, revealing systematic trade-offs between diversity and stability that shape task outcomes. Finally, we analyze the internal properties of both model families to probe their underlying mechanisms. These findings offer foundational insights into the probabilistic sampling behavior of LLMs and provide practical guidance on when to favor D- versus E-models. For web-scale applications—including recommendation, search, and conversational agents—our results inform model selection and configuration to balance diversity with reliability under real-world uncertainty.

\end{abstract}


%


\maketitle

\section{Introduction}

\begin{figure*}
  \includegraphics[width=\textwidth]{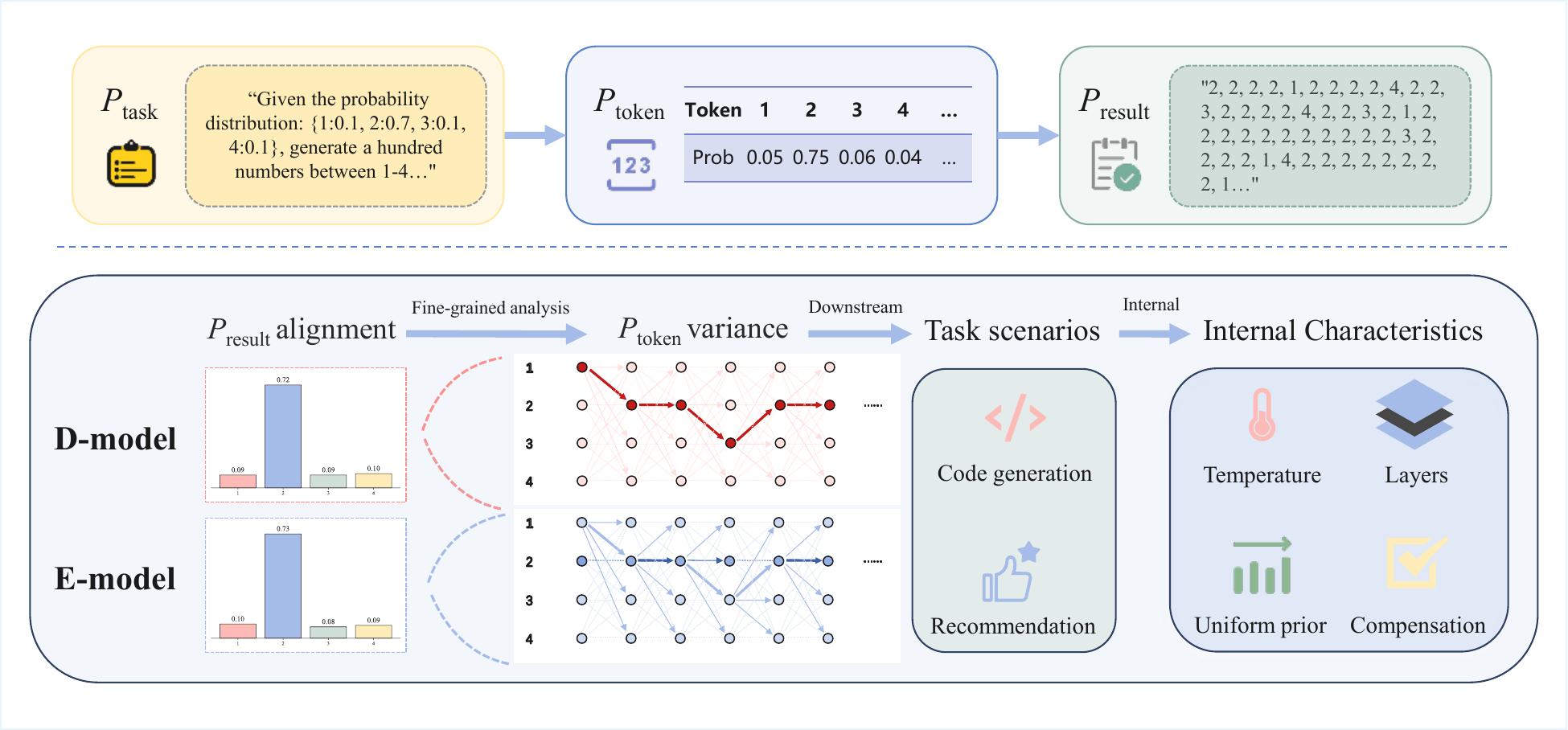}
  \caption{We analyze the probabilistic distribution sampling capabilities of LLMs by combining $P_\text{task}$, $P_\text{token}$, and $P_\text{result}$, and have verified the existence of both the D-model and the E-model. Although there are minor discrepancies in the $P_\text{result}$, which aligns somewhat with the $P_\text{task}$, they demonstrate distinct patterns at the fine-grained $P_\text{token}$ level. We have also analyzed both models within downstream task scenarios and their internal characteristics.}
  \label{fig:main}
\end{figure*}

\begin{figure}
  \includegraphics[width=0.47\textwidth]{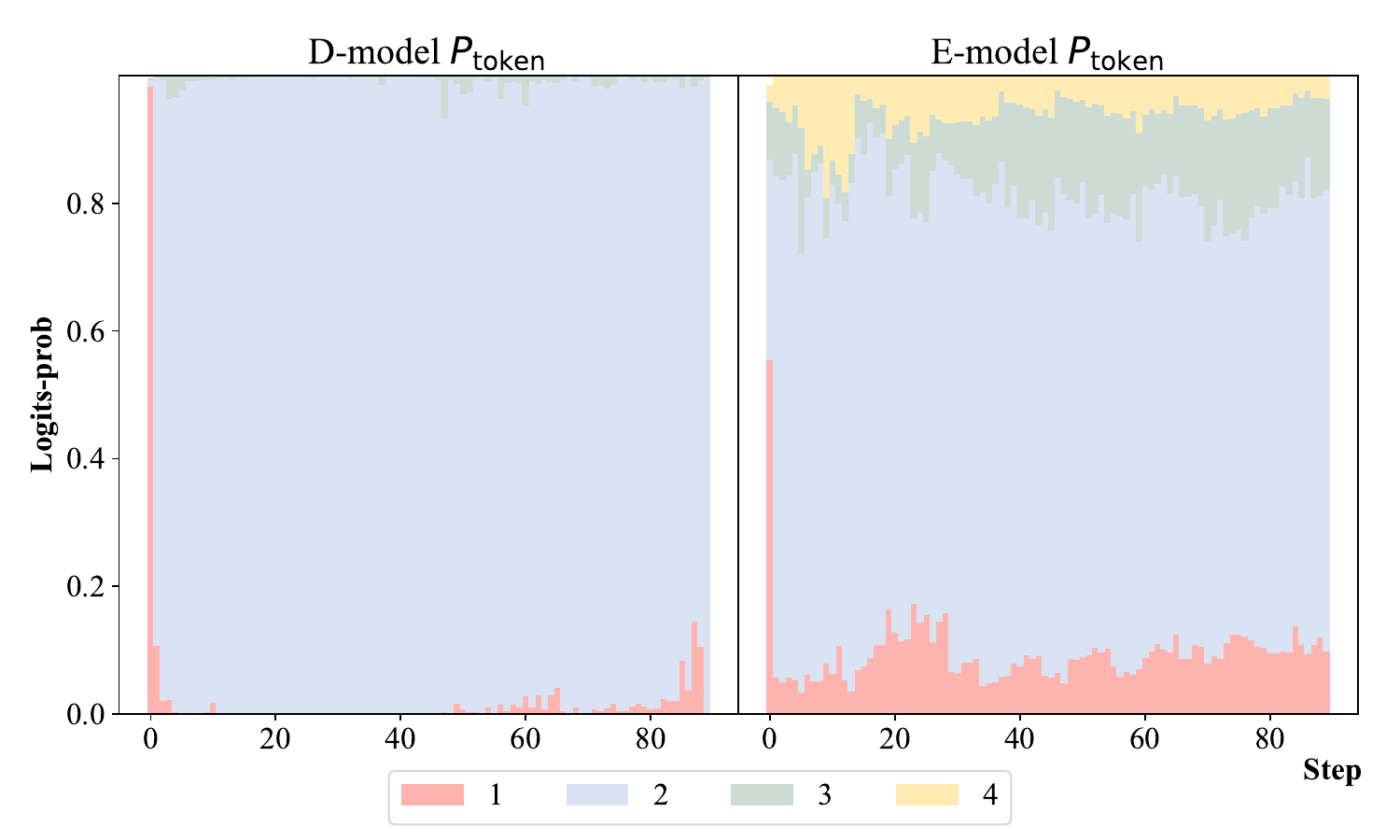}
  \caption{Comparison of $P_\text{token}$ between D- and E-model for distribution \{1: 0.1, 2: 0.7, 3: 0.1, 4: 0.1\}. The horizontal axis denotes the generation step, while the vertical axis represents the corresponding token probability distribution.}

  \label{fig:compare}
\end{figure}

The internet is increasingly relying on large language models (LLMs) to assist in code generation~\cite{maninger2025benchmarkingllmswebapi,johnson2025manipulatingllmwebagents} or moderate the exposure of information in search results, news feeds and recommendations~\cite{wang2025llmbasedrelevanceassessmentwebscale,dewan2025llmdrivenusefulnessjudgmentweb}. This moderation is not just a matter of convenience, but also a responsibility regarding the transparency of technology and algorithms, and user trust. When these models decide which content to display and how frequently, they are essentially sampling from a distribution of potential items, thereby shaping what users see, believe and do. Therefore, ensuring that this sampling process is auditable and explainable is a crucial aspect of the responsible development of internet technology.

The core foundation of LLMs' capabilities lies in their ability to model and learn the probability distributions embedded within large-scale text data. Their core mechanism can be summarized as follows: Given an existing text sequence (context), the model learns to predict the next most likely token. The model's core output is a probability distribution over the entire vocabulary, denoted as P(token | context). 
This distribution quantifies the likelihood of every possible next token given the current context. The model's generation process ultimately relies on sampling from this probability distribution, which we refer to as \textbf{$P_\text{token}$}.

Meanwhile, web tasks are inherently uncertain: In search, news feeds, and recommendations, a given user or query context induces a probability distribution over candidate items. When systems decide what to display and how frequently, they effectively sample from this distribution rather than selecting a single optimal item. Recommenders, for instance, aim not only to return the most probable items but also to balance diversity and uncertainty to meet heterogeneous user needs. We therefore model the task objective as a probability distribution over the candidate set, denoted \textbf{\(P_\text{task}\)}, which captures the intrinsic complexity of the web and provides a principled basis for reasoning under uncertainty.

This raises a critical question: \textbf{Can LLMs effectively understand and adapt to different task objective distributions?} In other words, can LLMs generate outputs aligned with the task-level objective distribution $P_\text{task}$ from their internal token-level probability distribution $P_\text{token}$? This question concerns not only the theoretical foundation of the model's generative capabilities but also its practical performance in complex tasks.


To explore this issue, this paper investigates the generative capabilities of LLMs from a probabilistic perspective. Prior work has examined their probability sampling behavior, including random number generation~\cite{vo2025bscoredetectingbiaseslarge, DBLP:journals/corr/abs-2505-00047,hopkins2023can} and coin flipping~\cite{DBLP:journals/corr/abs-2406-00092,DBLP:conf/acl/GuptaCGW0DC25,DBLP:journals/corr/abs-2506-09998}, revealing notable deficiencies~\cite{DBLP:conf/coling/GuPSC25,DBLP:journals/corr/abs-2506-09998,hopkins2023can,DBLP:conf/naacl/MeisterGH25}. However, these studies are largely empirical and scenario-specific, without systematically examining the mapping between the internal token-level distribution $P_\text{token}$ and the task-level objective distribution $P_\text{task}$, nor their practical implications in real-world tasks.


As shown in Figure~\ref{fig:main}, this study uses log probability to relate the token-level distribution $P_\text{token}$ produced by LLMs to the task-level target distribution $P_\text{task}$ and the resulting sampling distribution $P_\text{result}$, enabling a joint analysis of LLMs’ probability sampling capability.
During this analysis, we found that the $P_\text{token}$ of LLMs is not always characterized by extreme confidence~\cite{DBLP:journals/npjdm/GuDLY24,DBLP:journals/tacl/JiangADN21,DBLP:conf/naacl/MeisterGH25}, but rather exhibit two distinct and representative patterns. The first pattern exhibits deterministic global planning, whereby the probability of one token approaches 1 during generation while that of others approaches 0. However, this pattern does align with the target distribution overall. The second pattern demonstrates exploratory local planning and features a relatively uniform probability distribution that aligns more closely with the task probability distribution $P_\text{task}$.
Accordingly, we denote these two representative model types as \textbf{D-Models} (Deterministic Models) and \textbf{E-Models} (Exploratory Models), as illustrated in Figure~\ref{fig:compare}. In Section~\ref{sec:Validation}, we provide formal definitions and validation of these two model categories. Section~\ref{sec:task-Specific} then compares their performance and applicability in code generation and web-facing scenarios, particularly recommendation tasks, followed by an analysis of their internal characteristics in Section~\ref{sec:internal}.

The main contributions of this study are as follows:  
1) \textbf{Probabilistic view of web tasks:}
We formulate web tasks as probabilistic sampling processes with a task-level target distribution \(P_\text{task}\), and evaluate the alignment between \(P_\text{token}\), \(P_\text{task}\), and \(P_\text{result}\). Meanwhile, we identify and validate two model types—D (Deterministic) and E (Exploratory). 
2) \textbf{Task-specific comparison:}  
We evaluate both types across task-specific scenarios (e.g., code generation and recommendation), revealing a systematic diversity–stability trade-off and offering insights into when each model type performs favorably.  
3) \textbf{Internal mechanism analysis:}  
We further probe the internal mechanisms of both types, verifying their behavioral differences and uncovering the underlying causal factors.

\section{Model Types Validation}
\label{sec:Validation}
In this section, we first formalize the definitions of the D- and E-models. We then verify their existence through simulated distribution tasks with explicitly defined \( P_{\text{task}} \), comparing the alignment among \( P_{\text{token}} \), \( P_{\text{result}} \), and \( P_{\text{task}} \). Finally, we further validate the two models in a more practical setting.

\subsection{Definition of Models}
\label{definition}

Firstly, considering the discrete nature of the output vocabulary in LLMs, for a discrete random variable \( X  \in
\mathcal{X}\), we define two types of models based on the characteristics of their \( P_{\text{token}} \) and its alignment with \( P_{\text{task}} \). It should be noted that the two types of models do not differ significantly on the \( P_{\text{result}} \).

\textbf{D-model} generates a \( P_{\text{token}} \) that differs significantly from the \( P_{\text{task}} \), with one token assigned a substantially higher probability than all others. \textbf{E-model} is characterized by a \( P_{\text{token}} \) that closely aligns with the \( P_{\text{task}} \).
Formally, for each token generation step \( t \):

\begin{equation*}
\begin{aligned}
D\text{-model} \;\; &\triangleq\; 
\Big\{\,\text{LLM}\;\Big|\;\forall t,\;
P^t_{\text{token}}(X=x_i) \approx 
\begin{cases}
1.0, & \text{if } x_i = x_{\text{max}},\\
0.0, & \text{otherwise}.
\end{cases}
\Big\},
\\[6pt]
E\text{-model} \;\; &\triangleq\; 
\Big\{\,\text{LLM}\;\Big|\;\forall t,\;
P^t_{\text{token}}(X=x_i) \approx P_{\text{task}}(x_i), 
\quad \forall x_i \in \mathcal{X}
\Big\},
\end{aligned}
\end{equation*}

where \( x_{\text{max}} \) refers to the token with the highest probability. D-model results in a highly concentrated distribution with large fluctuations, leading to more diverse and unpredictable outputs.
E-model indicates that the model generates tokens in a stable manner, with minimal fluctuations in the token distribution, ensuring consistent outputs aligned with the task.

\subsection{Experimental Methods}
In this section, we first present a simulation experiment to obtain explicit $P_{\text{task}}, P_{\text{token}}$ and $P_{\text{result}}$. We then validate the two model types in practical tasks with experiments.

\subsubsection{Simulated Distribution Task}
\label{sec: simulated}

\paragraph{Distribution Definition}
Given the discrete nature of the output vocabulary of LLMs, the experimental tasks in this section are designed for discrete probability distribution sampling. This facilitates the alignment between the $P_{\text{task}}$, the $P_{\text{token}}$ and the $P_{\text{result}}$. The specific definitions of these three distributions are as follows:

First, $P_{\text{task}}$ refers to the target probability distribution that the task requires the model to sample from. For a discrete random variable $X \in \mathcal{X}$, its distribution is defined as:

\begin{equation}
  P_{\text{task}}(X = x_i) = p(x_i), \quad \text{where} \quad \sum_{x_i} p(x_i) = 1.
\end{equation}

In this experiment, we explicitly specify the task requirements and the target distribution $P_{\text{task}}(X)$ (where $X \in \mathcal{X}$) through the prompt, instructing the model to sample according to this distribution. The prompt template is:
\begin{tcolorbox}
\textbf{Simulated Tasks Prompt Template:} \\[4pt]
Given the probability distribution: $P_{\text{task}}(X)$, generate a number of hundreds in $\mathcal{X}$, and give the final list (numbers separated by commas) strictly according to the requirements, without adding any text or process description.
\end{tcolorbox}

Second, $P_{\text{token}}$ refers to the token-level probability distribution output by the LLM at each step during the generation process, based on the current context. Specifically, before generating the $t$-th token, the model outputs the raw logits $L^t(V)$ (where $V$ is the vocabulary, $\mathcal{X} \subset V$), which is converted into a probability distribution $P^t(V)$ via the Softmax function. The next token is then sampled from this distribution. We extract $P^t(X)$, corresponding to $X$, from $P^t(V)$ and normalize it to obtain $P_{\text{token}}^t(X)$ for that step:

\begin{equation}
    P_{\text{token}}^t(X = x_i) = \frac{P_t(x_i)}{\sum_{x' \in X} P_t(x')}.
\end{equation}

Third, $P_{\text{result}}$ refers to the empirical probability distribution presented by the number list parsed from the model's final output. Specifically, the sequence of numbers is extracted from the generated string, and the probability is approximated by frequency:
\begin{equation}
    P_{\text{result}}(X = x_i) = \frac{\text{count}(x_i)}{N},
\end{equation}

where $N$ is the length of the list.

\paragraph{Evaluation Metrics}
\label{sec:eval}
First, we define \textbf{e-score} to quantify the degree of extremeness in the model's $P_{\text{token}}^t$, reflecting the model's tendency to output concentrated distributions. It is computed as the average of the maximum probability value $p^t_{\text{max}}$ in the vocabulary probability distribution at each generation step:
$
    \text{e-score} = \frac{1}{T} \sum_{t=1}^{T} p^t_{\text{max}}.
$
This metric helps distinguish between D-Models (high e-scores) and E-Models (low e-scores).

Second, the Average Total Variation Distance (ATVD) measures the dispersion between distributions. Specifically, for any two distributions $P_{\text{A}}, P_{\text{B}} \subset {P_{\text{task}}, P_{\text{task}}, P_{\text{token}}}$, the formal definition of the ATVD between them is as follows:

\begin{equation}
    \text{ATVD}(P_{\text{A}}, P_{\text{B}}) = \frac{1}{n} \sum_{i=1}^{n} |P_{\text{A}}(X=x_i) - P_{\text{B}}(X=x_i)|,
\end{equation}

where $n$ is the number of possible values of $X$. 
Here, $P_{\text{token}} = \frac{1}{T} \sum_{t=1}^{T} P^t_{\text{token}}$, which represents the overall value.

Additionally, to measure the difference between $P_{\text{token}}$ and the $P_{\text{task}}$ at each finer-grained step, we further compute ATVD-step:
\begin{equation}
\begin{split}
    \text{ATVD-step} = \frac{1}{nT} \sum_{t=1}^{T}\sum_{i=1}^{n} |P^t_{\text{token}}(X=x_i) - P_{\text{task}}(X=x_i)|.
\end{split}
\end{equation}

\subsubsection{Practical Tasks}
\label{sec:mmlu}
To further validate the characteristics of D-model versus E-model, we analyzed the extremity of $P_\text{token}$ in tasks without explicitly specified probability distributions. Specifically, we conducted experiments on the MMLU benchmark, which comprises multiple categories of single-choice questions. By strictly prompting the LLMs to directly output options, we obtained the probability distribution corresponding to the first token generated by the LLM and recorded the e-score.

\subsection{Experimental Setups}
\subsubsection{Simulated Distribution Tasks}
To systematically evaluate the probability distribution sampling capability of LLMs, two types of representative distribution tasks are designed: \textbf{Extreme Distribution Task}: $\{1:0.1, 2:0.7, 3:0.1, 4:0.1\}$, exhibiting highly concentrated characteristics.
\textbf{Flat Distribution Task}: $\{1:0.1, 2:0.1, 3:0.1, 4:0.2, 5:0.1, 6:0.1, 7:0.1, 8:0.1, 9:0.1\}$, exhibiting approximately uniform characteristics.

For each model-task combination, we perform 10 independent sampling experiments. The final metrics are calculated as the average of these 10 runs to mitigate the impact of random fluctuations.

\subsubsection{Baseline Models}
\label{Models}
This study selects multiple LLMs as baseline models, covering both open-source and closed-source categories. 
\textbf{Open-source Models}: Llama-3.1-8B-Instruct~\cite{grattafiori2024llama3herdmodels}, Qwen-2.5-7B-Instruct~\cite{qwen2025qwen25technicalreport}, DeepSeek-R1-Distill-Qwen-7B~\cite{deepseekai2025deepseekr1incentivizingreasoningcapability}, Mistral-Small-24B-Instruct-2501~\footnote{\url{https://mistral.ai/news/mistral-small-3}}, Gemma-2-9B-IT~\cite{gemmateam2024gemma2improvingopen}, DeepSeek-v3~\cite{deepseekai2025deepseekv3technicalreport};
\textbf{Closed-source Models}: GPT-3.5-Turbo, GPT-4o\footnote{\url{https://chat.openai.com}}. Among them, the closed-source model and the DeepSeek-v3 are accessed via API.

It should be noted that due to API limitations, the API-accessed models only provide the logits values for the Top-5 tokens. Experimental observations indicate that the sum of the probability mass for these Top-5 tokens is close to 1.0, suggesting that the probabilities of the remaining tokens are negligible. Therefore, for elements in the target subset $X$ that do not appear in the Top-5, their logits values are uniformly set to negative infinity.

\subsubsection{Parameter Settings}
\label{Parameter}
To obtain raw log probabilities, we set the temperature parameter to 1.0 in these experiments. Unless specified otherwise, all subsequent experiments in this work were conducted at a temperature of 1.0. 

\subsection{Experimental Results}

\subsubsection{Sampling Performance}

\begin{table*}
  \caption{Sampling Performance Comparison of Models. Boldface indicates the best results under the corresponding metrics.}
  \label{tab:performance}
  \centering
  \small
\scalebox{0.95}{
  \begin{tabular}{ll|cccc|cccc}
    \toprule
    & Tasks & \multicolumn{4}{c}{Extreme distribution} & \multicolumn{4}{c}{Flat distribution} \\
    \cmidrule(lr){2-10}
    & Metrics & ATVD-step $\downarrow$ &\multicolumn{3}{c}{ATVD $\downarrow$} & ATVD-step $\downarrow$ & \multicolumn{3}{c}{ATVD $\downarrow$} \\
    \cmidrule(lr){2-2} \cmidrule(lr){4-6} \cmidrule(lr){3-3} \cmidrule(lr){8-10}\cmidrule(lr){7-7}
    Group & Model & Task/tok & Task/tok & Task/res  & Tok/res & Task/tok & Task/tok &  Task/res  & Tok/res \\
    \midrule
    \multirow{5}{*}{D-model} 
      & Llama-3.1 & 0.172 & $0.080 \pm 0.021$ & $0.080 \pm 0.022$ & $0.009 \pm 0.005$ & 0.193 & $0.026 \pm 0.012$ & $0.028 \pm 0.013$ & \textbf{$\textbf{0.003} \pm 0.005$} \\
      & Qwen-2.5  & 0.201 & $0.158 \pm 0.070$ & $0.159 \pm 0.070$ & \textbf{$\textbf{0.005} \pm 0.002$} & 0.151 & $0.023 \pm 0.009$ & $0.025 \pm 0.010$ & $0.014 \pm 0.008$ \\
      & DS-R1-Qwen-2.5 & 0.157 & $0.128 \pm 0.038$ & $0.131 \pm 0.041$ & $0.009 \pm 0.005$ & 0.165  & $0.031 \pm 0.020$ & $0.034 \pm 0.020$& $0.008 \pm 0.009$ \\
      & Deepseek-v3 & 0.189 & $0.080 \pm 0.059$ & $0.081 \pm 0.061$ & $0.005 \pm 0.004$ & 0.182 & \textbf{$\textbf{0.016} \pm 0.009$} & \textbf{$\textbf{0.011} \pm 0.007$} & $0.008 \pm 0.005$\\
      & AVG & 0.179 & $0.112 \pm 0.047$ & $0.113 \pm 0.049$ & $0.007 \pm 0.004$ & 0.173  & $0.024 \pm 0.013$ & $0.024 \pm 0.013$& $0.008 \pm 0.007$ \\
    \midrule
    \multirow{5}{*}{E-model} 
      & Mistral-Small & \textbf{0.090} & $0.084 \pm 0.027$ & $0.092 \pm 0.031$ & $0.018 \pm 0.006$ & \textbf{0.048} & $0.022 \pm 0.016$ & $0.036 \pm 0.016$ & $0.023 \pm 0.010$ \\
      & Gemma-2 & \textbf{0.090} & \textbf{$\textbf{0.036} \pm 0.017$} & \textbf{$\textbf{0.042} \pm 0.026$} & $0.024                         \pm 0.014$ & 0.131 & $0.024 \pm 0.004$ & $0.021 \pm 0.003$ & $0.016 \pm 0.006$ \\
      & GPT-3.5-Turbo & 0.099 & $0.098 \pm 0.018$ & $0.104 \pm 0.024$ & $0.015 \pm 0.009$ & 0.094 & $0.082 \pm 0.010$ & $0.055 \pm 0.023$ & $0.040 \pm 0.011$ \\
      & GPT-4o & 0.116 & $0.082 \pm 0.024$ & $0.089 \pm 0.026$ & $0.016 \pm 0.007$ & 0.106 & $0.031 \pm 0.012$ & $0.013 \pm 0.007$ & $0.035 \pm 0.016$ \\
      & AVG & 0.101 & $0.075 \pm 0.022$ & $0.082 \pm 0.027$  & $0.018 \pm 0.009$ & 0.095 & $0.040 \pm 0.011$ & $0.031 \pm 0.012$ & $0.029 \pm 0.011$ \\
    \bottomrule
  \end{tabular}
  }
\end{table*}

\begin{figure}
    \centering 
    \includegraphics[width=0.47\textwidth]{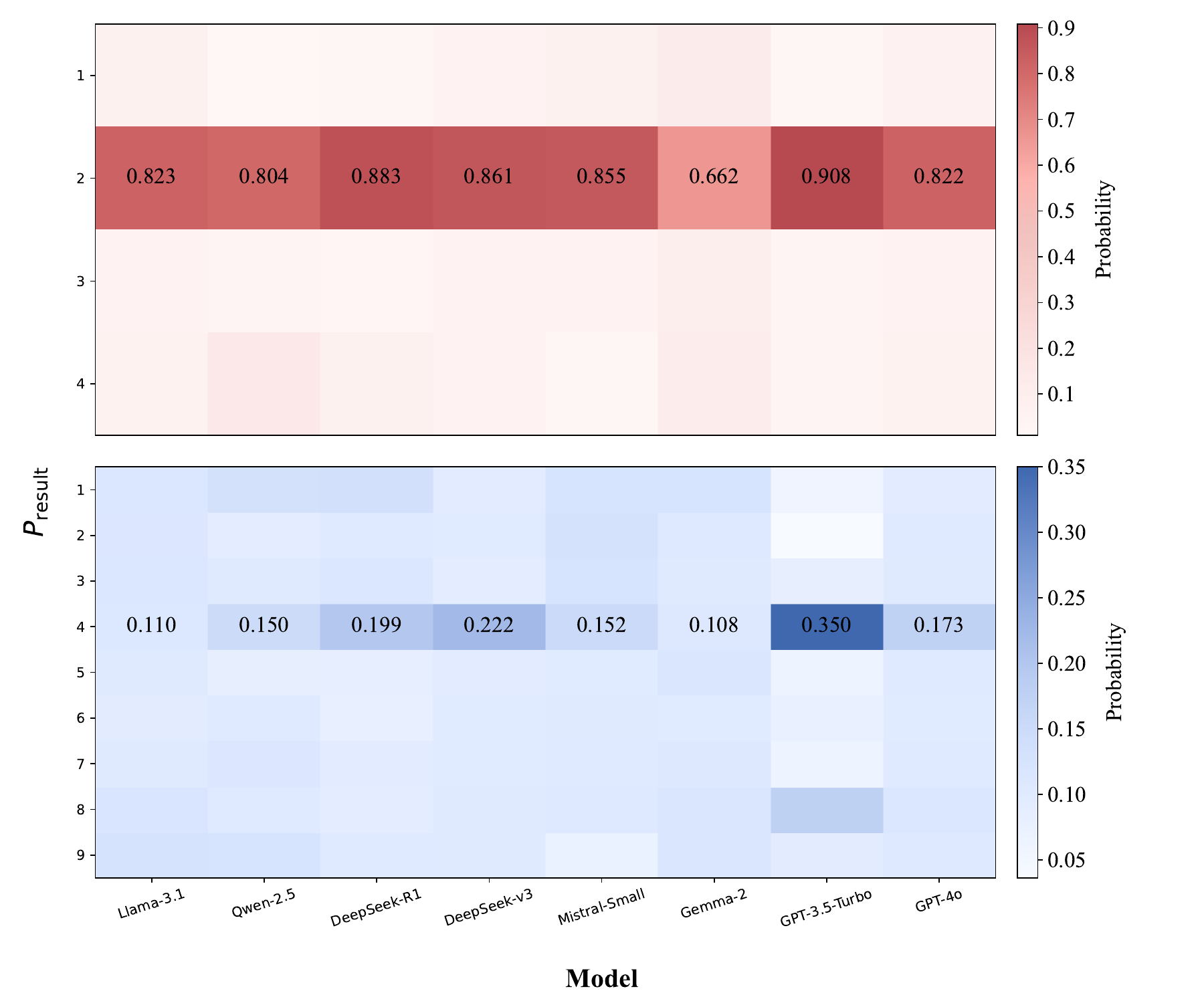}
    \caption{The $P_\text{result}$ of various models.
    In the extreme task (top), $P_{\text{result}}(2)$ > $P_{\text{task}}(2) = 0.7$, whereas in the flat task (bottom), $P_{\text{result}}(4)$ > $P_{\text{task}}(4) = 0.2$ for most models.
    }
    \label{fig:logits}
\end{figure}

\textbf{The overall alignment of $P_{\text{token}}$ and $P_{\text{result}}$ with $P_{\text{task}}$ is limited, as shown in Table~\ref{tab:performance}.} In the extreme distribution task, in particular, the ATVD$(P_\text{task}, P_\text{result})$ and ATVD$(P_\text{task}, P_\text{token})$ concentrate near 0.10, implying an average probability gap of about 0.10 per x. This reveals that while current LLMs can sample simple distributions, they struggle to model distributions precisely. Conversely, in flat distribution task, the models' ATVD values are consistently lower, indicating better sampling performance for more uniform distributions. In short, LLMs exhibit greater uncertainty and bias when the target distribution is extreme, but align more reliably when it is nearly uniform.

The model's internal probability distribution closely aligns with its sampling behavior. As shown in Table~\ref{tab:performance}, $\text{ATVD}(P_{\text{token}}, P_{\text{result}})$ remains below 0.04 across all models; for example, Qwen-2.5 reaches 0.005 in the extreme distribution task. This consistency indicates that $P_{\text{token}}$ effectively drives $P_{\text{result}}$, reflecting a strong synergistic relationship between them.

Additionally, as shown in Figure~\ref{fig:logits}, we observe a clear monotonic pattern: tokens \(x\) with higher \(P_{\text{task}}(x)\) tend to receive higher  \(P_{\text{token}}(x)\) and appear higher \(P_{\text{result}}(x)\); conversely, tokens with lower \(P_{\text{task}}(x)\) receive lower \(P_{\text{token}}(x)\) and \(P_{\text{result}}(x)\). We hypothesize that LLMs track relative probabilities via natural language understanding, rather than mapping absolute magnitudes faithfully, leading to systematic overestimation of high-probability items and underestimation of low-probability ones. Moreover, for nearly all models we find \(\text{ATVD-step} > \text{ATVD}\!\left(P_{\text{task}}, P_{\text{token}}\right)\), highlighting a finer-grained mismatch between \(P_{\text{token}}\) and \(P_{\text{task}}\) at the step level.

\subsubsection{Two Patterns Analysis}

\begin{figure}
  \centering
  \includegraphics[width=\linewidth]{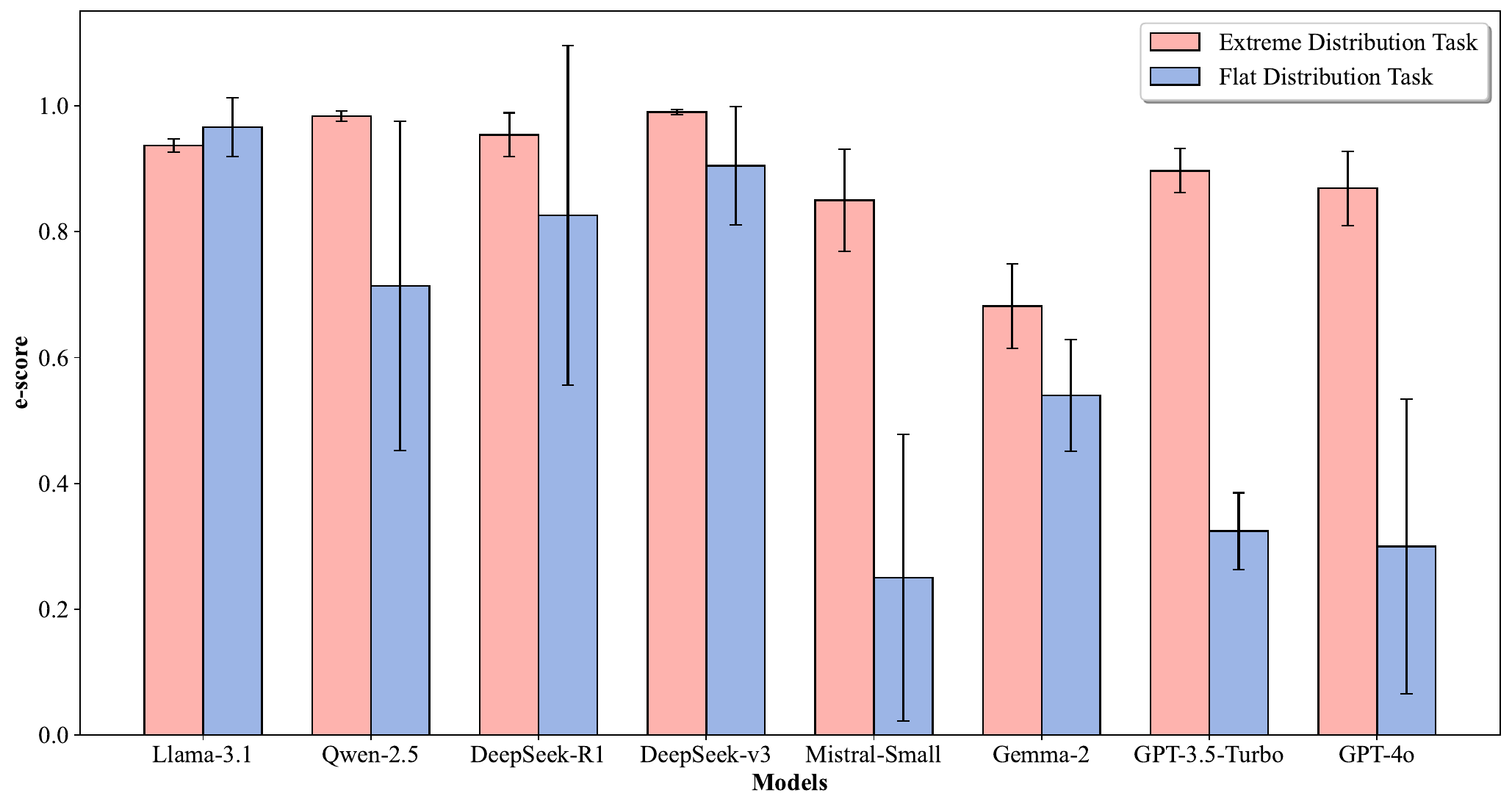}
  \caption{E-scores comparison of models, with error bars representing standard deviations.}
  \Description{}
  \label{escore}
\end{figure}

\textbf{In probability distribution sampling tasks, LLMs show two distinct patterns as shown in Figure~\ref{escore}, especially in the flat distribution task.} Based on these patterns, we categorize them into two model types as defined in Section~\ref{definition}.
\textbf{D-models} (Llama-3.1, Qwen-2.5, DS-R1-Qwen-2.5, and DeepSeek-v3) consistently exhibit highly concentrated distributions, with e-scores generally above 0.9.  
This indicates extreme confidence, where the maximum probability approaches 1.0 regardless of task, revealing a tendency toward pattern fixation even when a uniform output is required. 
\textbf{E-models} (Mistral-Small, Gemma-2, GPT-3.5-Turbo, GPT-4o) display greater task-dependent adaptability. In the extreme distribution task, where the maximum probability in $P_\text{task}$ is 0.7, E-models exhibit relatively high e-scores (e.g., GPT-3.5-Turbo reaches 0.897), though still lower than D-models. Conversely, in the flat distribution task, where the maximum probability in $P_\text{task}$ is only 0.2, they show significantly lower e-scores. 

Crucially, although E-models exhibit lower e-scores than D-models and show some adaptation to \(P_\text{task}\), all their e-scores remain higher than the maximum probability in the target \(P_\text{task}\). This suggests that while E-models can adjust distribution flatness according to the task, they still display a consistent bias, producing probability distributions more concentrated than required and failing to achieve full sampling fidelity.

The two model types also differ across task settings: E-models perform better in extreme-distribution scenarios, whereas D-models excel in flat-distribution ones. As shown in Table~\ref{tab:performance}, in the extreme distribution task D-models yield higher average ATVD\((P_\text{task}, P_\text{result})\) (0.113) than E-models (0.082), while in the flat distribution task D-models reach lower ATVD values (0.024) than E-models (0.031).

Analysis of ATVD-step further indicates that D-models exhibit larger stepwise fluctuations, with generated tokens deviating more from \(P_\text{task}\) (e.g., Qwen-2.5 shows 0.201). In contrast, E-models show ATVD-step values consistent with ATVD(task, token), reflecting higher intra-step stability. Details of the \(P_\text{token}\) evolution during generation are provided in Appendix~\ref{app:logits}.

\subsubsection{Further Validation of Model Classification}

\begin{figure}
  \centering
  \includegraphics[width=\linewidth]{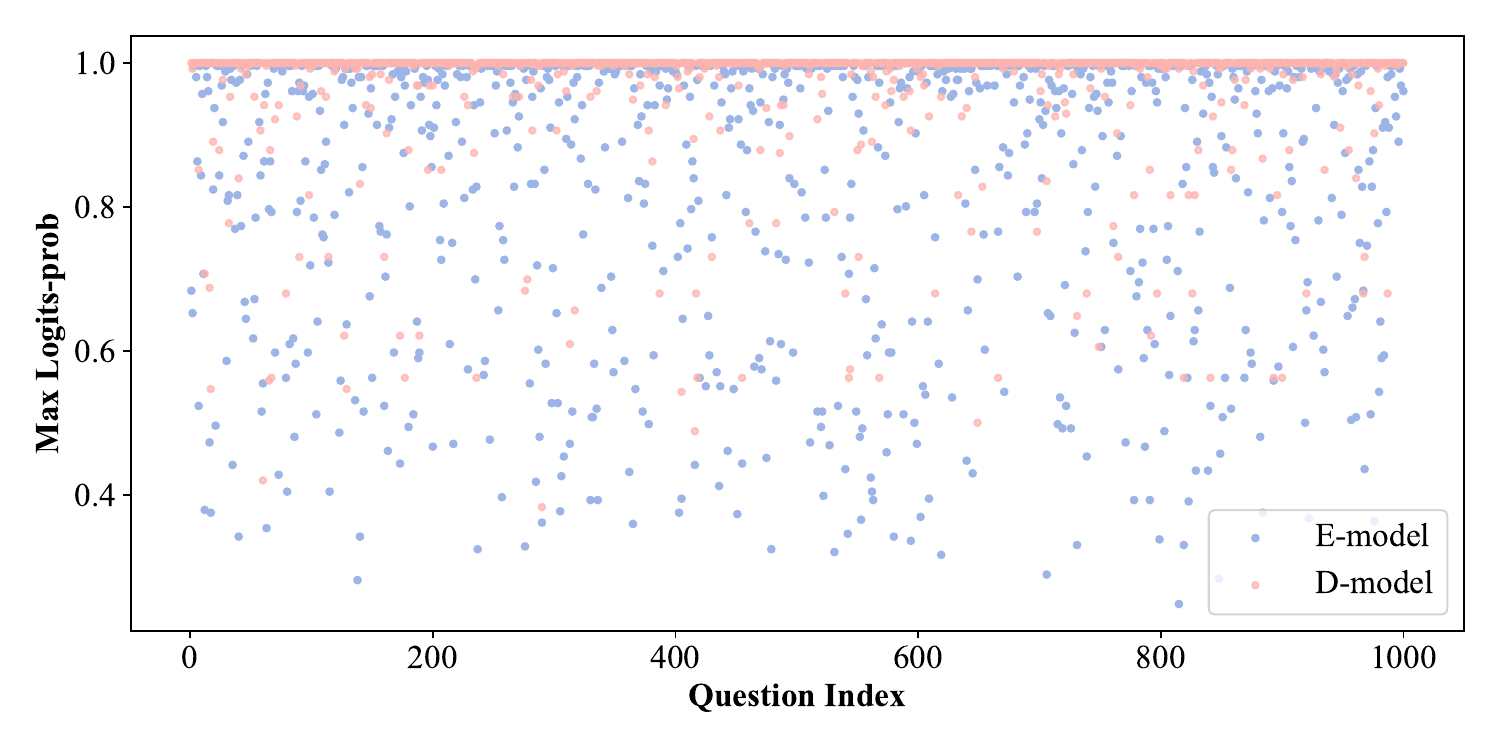}
  \caption{The distribution of maximum logits probabilities for 1,000 random problems in MMLU dataset, as demonstrated by D-model Qwen-2.5 and E-model Mistral-Small.}
  \Description{}
  \label{fig:mmlu}
\end{figure}

The experimental results are shown in the Figure~\ref{fig:mmlu}, for e-score results of additional models, see the Appendix~\ref{app:mmlu}. During generation, the probability mass of the maximum-probability token under the D-model remains concentrated near the top, whereas under the E-model it is more dispersed at comparatively lower values. This demonstrates that, for real-world tasks, the D-model exhibits highly confident token probabilities and a relatively concentrated $P_\text{token}$, while the E-model generates more conservative $P_\text{token}$ based on task content.

\section{Task-Specific Properties Comparison}
\label{sec:task-Specific}

\subsection{Experimental Methods}
In this experiment, we propose that LLMs implicitly model task-specific probability distributions when processing particular tasks. For a given context, there exists an underlying task-level distribution \(P_{\text{task}}\), from which LLMs sample to produce the result distribution \(P_{\text{result}} \sim P(\text{result} \mid \text{context})\). Concretely, the model predicts tokens autoregressively by sampling from the token-level distribution \(P_{\text{token}} \sim P(\text{token} \mid \text{context})\). This process relies on the model's ability to model the probability distribution at each step of the generation process. We then analyze the performance of two model categories (D- and E-models) across distinct task scenarios to examine their effectiveness in task-specific applications.

\subsubsection{Code Generation}
\paragraph{Distribution Definition}

The objective of code generation is to infer an appropriate function implementation from a given context. We model this process as a conditional probability distribution, $P_\text{task} \sim P(\text{code} \mid \text{problem})$, indicating that, given the problem description, the model seeks to generate code samples with high probability under this distribution. 

\paragraph{Evaluation Methodology}
To quantify the diversity of model responses, except for the response's vector cosine similarity, we combine the traditional pass@5 and pass@1 metrics with an exponential scaling that accounts for the increasing difficulty of improvement at higher accuracy levels. We define $\Delta pass = e^{\text{pass@5}} - e^{\text{pass@1}}$.
A larger $\Delta pass$ indicates that the model explores a broader solution space and generates more diverse candidate answers.

\subsubsection{Recommendation}
\label{sec:recomm}
\paragraph{Distribution Definition}

The goal of recommendations is to predict the items most likely to interest a user given their profile and interaction history. We formulate this as a conditional probability distribution, \( P_\text{task} \sim P(\text{items} \mid \text{profile}, \text{history}) \), where the model estimates item probabilities and selects the top-\(k\) recommendations.

Candidate items are first obtained using UCF (User-based Collaborative Filtering) and ICF (Item-based Collaborative Filtering) to generate \(n\) candidates. The LLM is then prompted to recommend items based on the user information and these candidates, following the prompt template detailed in Appendix~\ref{app:prompt}.

\paragraph{Evaluation Metrics}
We select candidate items generated by UCF and ICF methods, and task the model with choosing the top-$k$ items for recommendation. We use precision to evaluate the quality of the recommendations' rankings. The can-hit rate reflects the model's recommendation divergence by measuring the proportion of predicted recommended items that appear in the candidate list.

\subsection{Experimental Setups}
\paragraph{Datasets}
We use the HumanEval benchmark~\cite{chen2021codex} for the code task. For the recommendation, we use the MovieLens-1M dataset~\cite{10.1145/2827872}, which includes user rating data and item information.

\paragraph{Hyperparameters}
In the recommendation task, both UCF and ICF each generate 10 candidates. The LLM generates \(k=10\) recommendations. Temperature settings follow Section~\ref{Parameter}.

\paragraph{Models}

The baseline models used in this section are the same as those described in Section~\ref{Models}. 
In the code generation, we employ Qwen3-Embedding-4B~\cite{qwen3embedding} 
to convert the generated text into embedding vectors and to compute the similarity.

\subsection{Experimental Results}

\begin{table}
  \caption{Performance for Code and Recommendation Tasks. The DS-R1-Qwen-2.5 lacks 
$\Delta pass$ as it rarely produced complete outputs even with a greatly increased max-token limit.}
  \label{tab:performance_task}
  \centering
    \small
  \begin{tabular}{ll|cc|cc}

    \toprule
    & Tasks & \multicolumn{2}{c}{Code} & \multicolumn{2}{c}{Recommendation} \\
    \cmidrule(lr){2-6}
    Group & Model & $\Delta pass$ & sim & Prec & Can-hit \\
    \midrule
    \multirow{5}{*}{D-model} 
      & Llama-3.1 & 0.444 & 0.941 & 0.172 &  0.231 \\
      & Qwen-2.5 & 0.368 & 0.943 & 0.138 &  0.175 \\
      & DS-R1-Qwen-2.5 & - & 0.945 & 0.296 &  0.421 \\
      & Deepseek-v3 & 0.080 & 0.981 & 0.350  & 0.500 \\
      & AVG & 0.297 & 0.952 & 0.179  & 0.475 \\
    \midrule
    \multirow{4}{*}{E-model}
      & Mistral-Small & 0.272 & 0.783 & 0.320  & 0.466 \\
      & Gemma-2 & 0.209 & 0.929 & 0.343 & 0.482 \\
      & GPT-3.5-Turbo & 0.351 & 0.912 & 0.348 & 0.496 \\
      & GPT-4o & 0.183 & 0.942 & 0.341 & 0.499 \\
      & AVG & 0.264 & 0.893 & 0.337 & 0.491 \\
    \bottomrule
  \end{tabular}
\end{table}

It is evident from the experimental results that the D-model and the E-model demonstrate disparate performance in code and recommendation evaluations. This observation is indicative of the presence of discrepancies in their generative capabilities and task adaptability. The D-model exhibits stronger deterministic planning, yet it tends to deviate from target distributions and constraints during generation. Conversely, the E-model maintains better constraint adherence during generation, although it may engender unnecessary diversity overall.

In the code generation task, as illustrated in the Table~\ref{tab:performance_task}, the code fragments generated by the D-model through multiple passes demonstrate a higher degree of similarity among themselves and achieve a higher $\Delta pass$ value when compared to the E-model. This mechanism is rooted in the D-model's deterministic planning, wherein $P_\text{token}$ exhibits significant concentration at each step. The decoding process entails the introduction of minor perturbations to the primary code, while preserving the fundamental deterministic content. These minor adjustments enable successive passes to refine details in the solution, leading to higher values for the $\Delta pass$. In contrast, the E-model’s flatter 
$P_\text{token}$ leads to broader exploration, but in syntax-sensitive code tasks, global changes more easily introduce semantic or execution errors, yielding lower $\Delta\text{pass}$.

In the recommendation task, as illustrated in the Table~\ref{tab:performance_task}, the D-model demonstrates a substantially diminished can-hit in comparison to the E-model. This finding signifies a pronounced inclination towards generating a more diverse array of entries beyond the confines of the candidate list, accompanied by a notable deviation from the task's fundamental requirement of selecting from candidates. Conversely, the E-model demonstrates a higher can-hit value, suggesting that its recommendations are more likely to select items from the candidate list. The outcomes of the E-model have been found to be more concentrated and accurate, demonstrating a more stable alignment with $P_\text{task}$.

Overall, for deterministic tasks (e.g., mathematics, code generation), the D-model has been demonstrated to generate deterministic solutions and refine minor errors through multiple iterations. Conversely, the E-model's task alignment during generation introduces greater uncertainty that may affect performance. However, in tasks involving explicit candidates (e.g., recommendation, search), the D-model's global planning may deviate confidently from candidates during generation, while the E-model’s alignment during generation yields more accurate results.

\section{Internal Characteristics Analysis}
\label{sec:internal}
In this section, we analyze the internal characteristics of the two model types based on simulation experiments in Section~\ref{sec: simulated}. 
\subsection{Temperature}
\label{temperature}

\begin{figure}
    \centering 
    \includegraphics[width=0.47\textwidth]{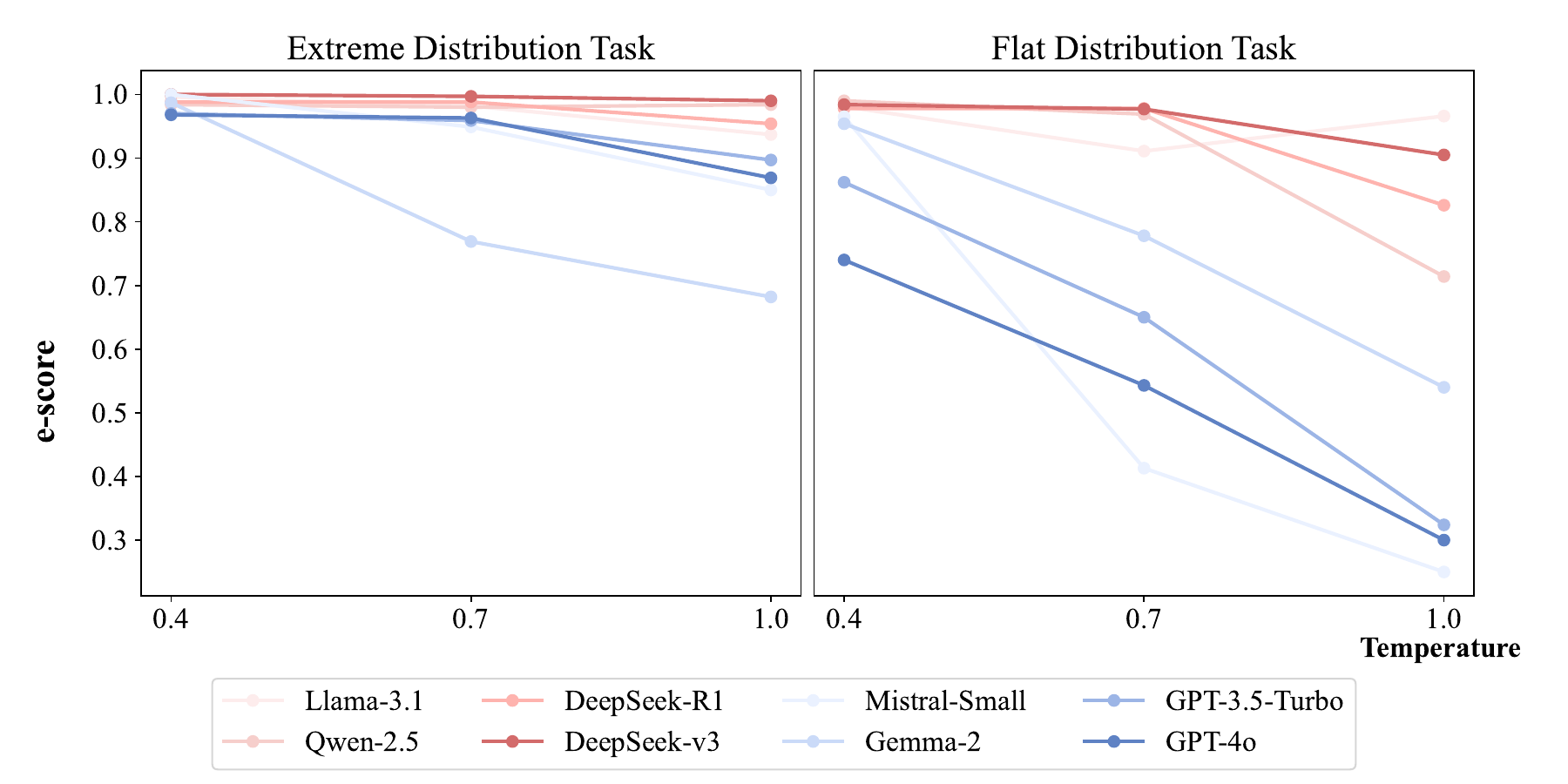}
    \caption{The e-scores of various models change with temperature. Among them, the first four models in the legend are D-models, while the last four are E-models.}
    \label{fig:temperature}
\end{figure}

By varying the temperature parameter, we examined its effect on the extremity of \(P_{\text{token}}\) in model outputs. 
Results indicate that E-models are more temperature-sensitive and exhibit stronger controllability, allowing flexible adaptation across tasks, while D-models tend to maintain low-entropy sampling.

As shown in Figure~\ref{fig:temperature}, increasing temperature leads to lower e-scores for all models, reflecting flatter probability distributions and reduced extremity. However, the decline in e-score for D-models is notably smaller than that for E-models.

This pattern aligns with temperature’s role in reshaping the original distribution through scaling. For D-models, even under higher temperatures, \(P_{\text{token}}\) remains sharply concentrated, suggesting that the model consistently leans toward a high-confidence option in token prediction, and temperature adjustments can hardly significantly alter its extremely confident behavior. In contrast, E-models display greater sensitivity, as their flatter baseline distributions enable more effective temperature-based control. Temperature can be lowered to enhance certainty in deterministic tasks, while it can be raised to promote diverse generation in creative tasks.

\subsection{Prior Analysis}

\begin{figure}
    \centering 
    \includegraphics[width=0.47\textwidth]{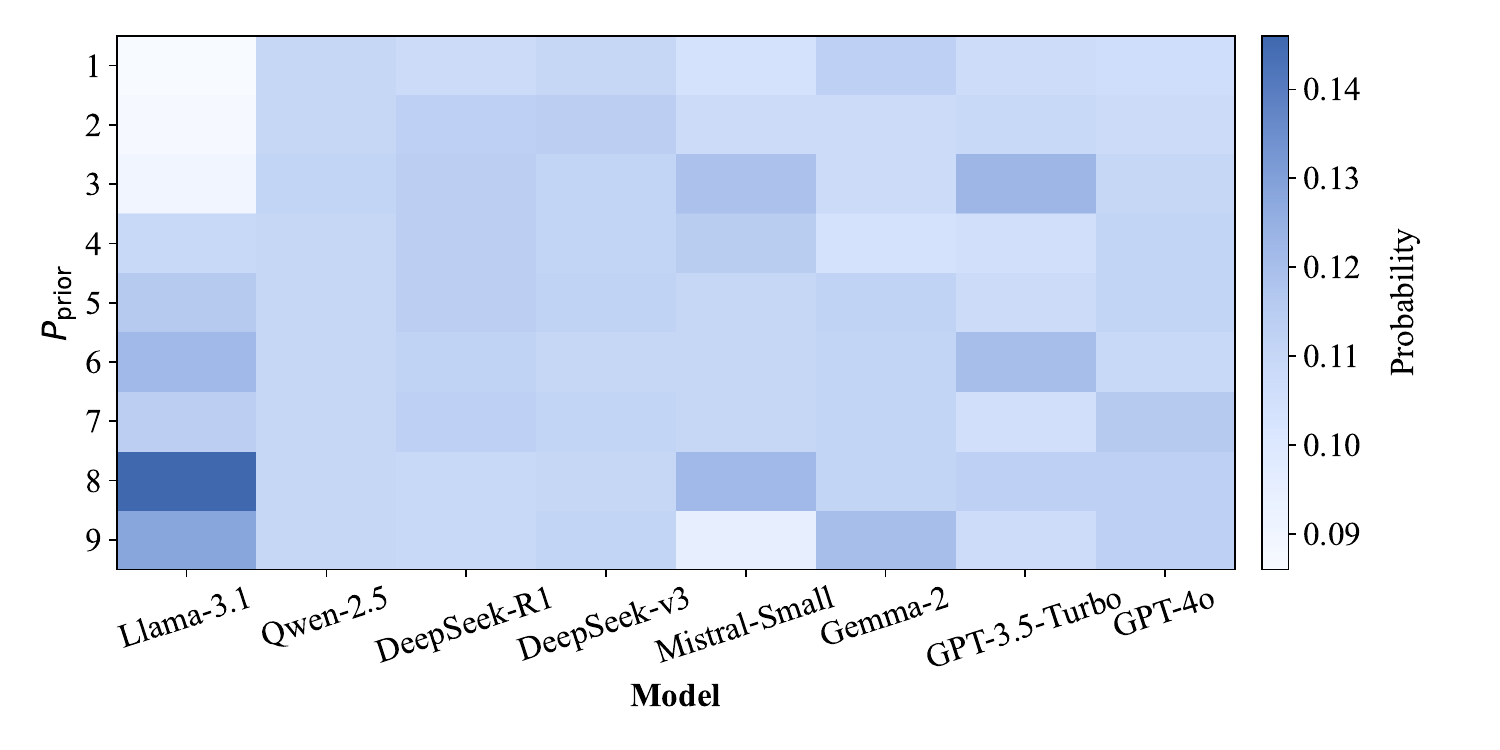}
    \caption{Verify the prior distribution $P_\text{result}$ for each model. For the same model, the closer the colours of the different numbers are, the more uniform the prior distribution.}
    \label{fig:prior}
\end{figure}

\begin{table}
  \caption{Prior Validation Across Models}
  \label{tab:prior}
  \centering
  \small
  \scalebox{0.91}{
  \begin{tabular}{ll|ccc}
    \toprule
     &  & \multicolumn{3}{c}{ATVD $\downarrow$} \\
    Group & Model & Task/res & Task/tok & Tok/res \\
    \midrule
    \multirow{5}{*}{D-model}
      & Llama-3.1 & $0.021 \pm 0.016$ & $0.024 \pm 0.011$ & $0.017 \pm 0.008$ \\
      & Qwen-2.5  & $0.001 \pm 0.000$ & $0.002 \pm 0.001$ & $0.001 \pm 0.001$ \\
      & DS-R1-Qwen-2.5 & $0.005 \pm 0.004$ & $0.008 \pm 0.004$ & $0.009 \pm 0.004$ \\
      & DeepSeek-v3 & $0.002 \pm 0.000$ & $0.008 \pm 0.000$ & $0.008 \pm 0.000$ \\
      & AVG & $0.007\pm 0.005$ & $0.011\pm 0.004$ & $0.009\pm 0.003$ \\
    \midrule
    \multirow{5}{*}{E-model}
      & Mistral-Small & $0.017 \pm 0.007$ & $0.012 \pm 0.005$ & $0.026 \pm 0.004$ \\
      & Gemma-2 & $0.005 \pm 0.003$ & $0.020 \pm 0.005$ & $0.022 \pm 0.007$ \\
      & GPT-3.5-Turbo & $0.014 \pm 0.013$ & $0.035 \pm 0.022$ & $0.042 \pm 0.028$ \\
      & GPT-4o & $0.004 \pm 0.002$ & $0.016 \pm 0.005$ & $0.018 \pm 0.005$ \\
      & AVG & $0.010\pm 0.006$ & $0.021\pm 0.009$ & $0.027\pm 0.011$ \\
    \bottomrule
  \end{tabular}
  }
\end{table}

To eliminate potential biases arising from intrinsic numeric preferences that could affect the resulting probability distributions, we analyze the prior probabilities over the possible values \(x_i\) of the discrete random variable \(X\). The results indicate that the tested LLMs exhibit no significant prior bias toward any particular \(x_i\).

Specifically, without specifying any explicit target distribution, we ask the LLMs to randomly generate samples of the target random variable \(X \in \mathcal{X}=\{x_i\}\), yielding an empirical prior distribution \(\hat{P}_{\text{prior}}\). The prompt template is detailed in the Appendix~\ref{app:prompt}. 

We then compute the ATVD as in Section~\ref{sec:eval}. As shown in the Tables~\ref{tab:prior} and Figure~\ref{fig:prior}, the ATVD values are generally smaller than in the simulated-distribution setting, and \(\hat{P}_{\text{prior}}\) is highly similar to the uniform distribution \(U(\mathcal{X})\) except for Llama-3.1. 

This indicates that, under our setups, the LLM’s prior bias over \(\mathcal{X}\) is negligible; therefore, the deviations from \(P_{\text{task}}\) observed previously are more likely attributable to the sampling mechanism along \(P_{\text{token}}\!\to\! P_{\text{result}}\), rather than to any inherent preference for particular numbers.

\subsection{Layer Probability Detection}

\begin{figure*}
    \centering 
    \includegraphics[width=\textwidth]{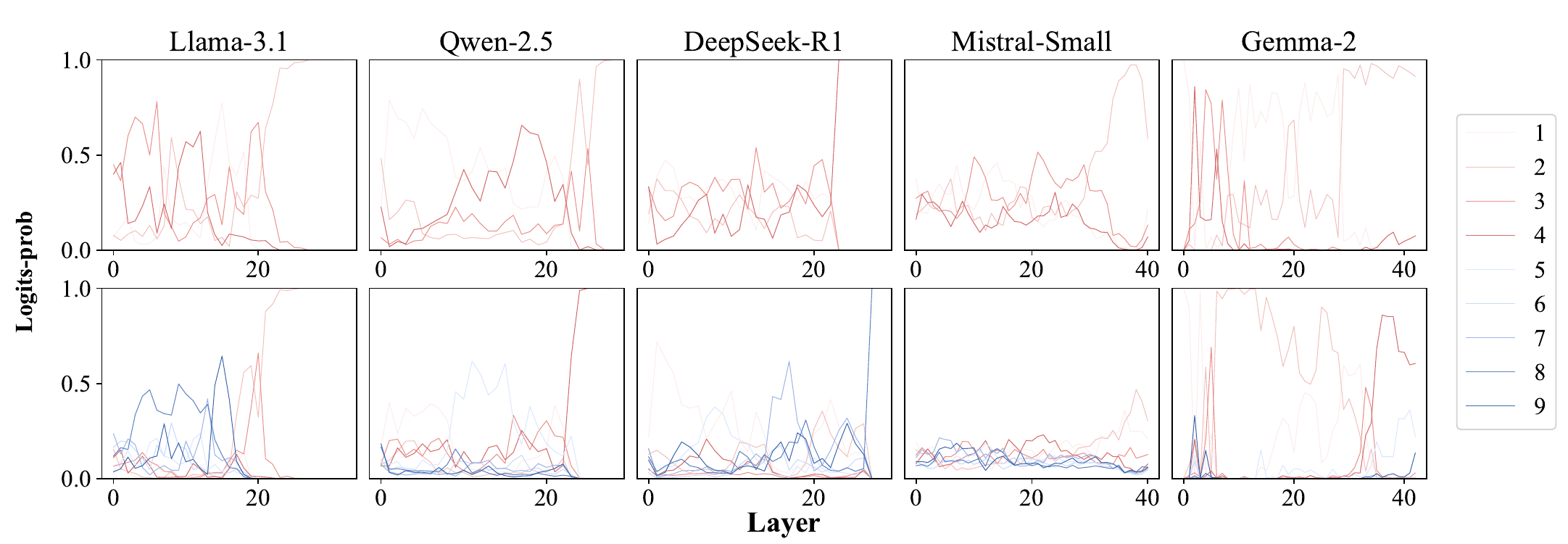}

    \caption{Detection of intermediate-layer probability evolution. Columns correspond to Llama-3.1, Qwen-2.5, DS-R1-Qwen-2.5, Mistral-Small, and Gemma-2. Row 1: Extreme distribution task results; Row 2: Flat distribution task results.}
    \label{fig:layer}
\end{figure*}


To further analyze the generation mechanisms of D-Models and E-Models, we track the layer-wise evolution of \(P_{\text{token}}\) across model layers. 

Specifically, using an intervention-based activation extraction method, we sampled hidden states from each layer and reconstructed layer-wise logits through the final layer’s normalization and projection mechanisms. By applying the Softmax to target token values, we obtained intermediate-layer distributions \(P^l_{\text{token}}\), allowing visualization and quantitative analysis of their evolution.


Due to API limitations, experiments were conducted on open-source models only, except for DeepSeek-v3. Figure~\ref{fig:layer} shows layer-wise probability changes under extreme and flat tasks. Except for Gemma-2, both D- and E-Models display relatively uniform probabilities in early layers that sharply concentrate near the output, referred to as the \emph{up-layer}, where target semantics are integrated and the dominant token is determined.


Beyond the up-layer, the two model types diverge. In extreme tasks, the D-Model maintains nearly deterministic probabilities (close to 1.0), indicating strong confidence, while the E-Model gradually converges toward the task distribution \(P_{\text{task}}\). In flat tasks, the D-Model still deviates from \(P_{\text{task}}\), whereas the E-Model aligns 
\(P^l_{\text{token}}\) more closely with \(P_{\text{task}}\), demonstrating better distribution calibration.

In summary, E-Models demonstrate stronger probability convergence and alignment with \(P_{\text{task}}\), enabling dynamic adjustment of token probabilities according to task goals. Conversely, D-Models tend to preserve extreme confidence, leading to overly deterministic outputs and reduced flexibility in tasks requiring diversity.

\subsection{Quota Compensation Analysis}

\begin{figure}
    \centering 
    \includegraphics[width=0.47\textwidth]{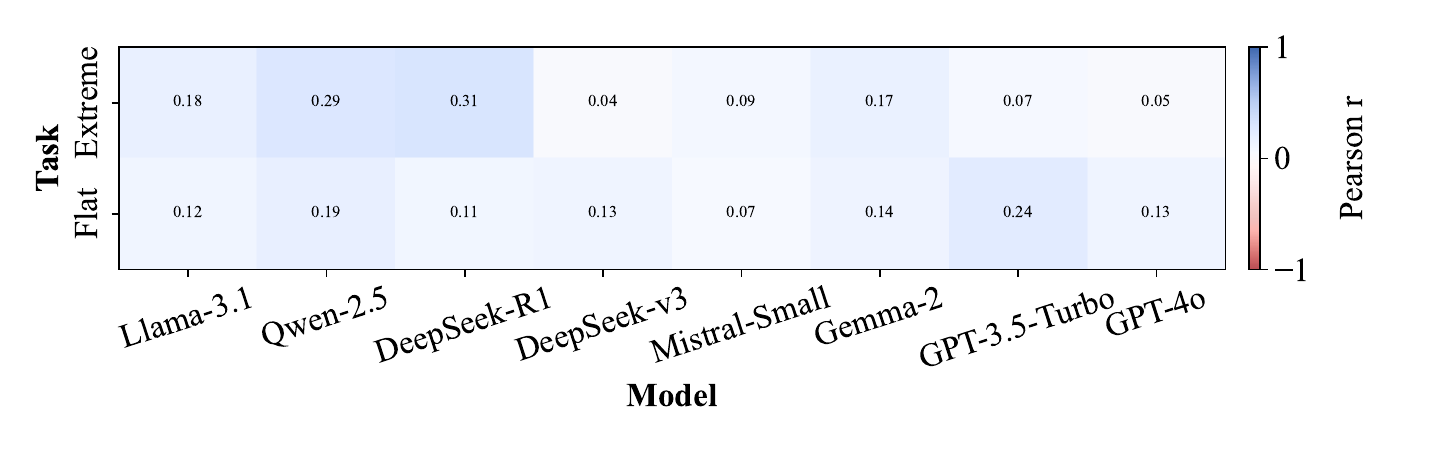}
    \caption{Pearson correlation coefficients between \(d_i^{t}\) and \(\Delta P_{\text{token}}^{t}(X = x_i)\) across models on extreme and flat tasks.}
    \label{fig:Quota}
\end{figure}

In this section, we investigate how, in the D-model, the final sampling distribution can resemble that of the E-model even when \(P_{\text{token}}\) is extremely peaked. To explain this phenomenon, we hypothesize that the D-model’s sampling follows a quota-compensation mechanism. The findings suggest that this hypothesis is invalid, thereby indicating that the D-model is more likely to adopt a deterministic global planning approach.

At generation step \(t\), let the empirical distribution of the already generated sequence be \(P_{\text{result}}^{t}\). For any token \(x_i \in \mathcal{X}\), if there is a large discrepancy between \(P_{\text{result}}^{t}(X=x_i)\) and the target distribution \(P_{\text{task}}(X=x_i)\), we posit that the model compensates by increasing the logit of \(x_i\) when generating the next token, thereby raising \(P_{\text{token}}^{t+1}(X=x_i)\) and manifesting the E-model’s extremeness.

We define the residual
\(
d_i^{t} \;=\; P_{\text{task}}(X = x_i)\;-\;P_{\text{result}}^{t}(X = x_i).
\)
If \(|d_i^{t}|\) is large, i.e., the current token frequency departs substantially from the target, we assume a probability adjustment:
\[
\Delta P_{\text{token}}^{t}(X = x_i)
\;=\; P_{\text{token}}^{t+1}(X = x_i) - P_{\text{token}}^{t}(X = x_i)
\;=\; \lambda \cdot d_i^{t},
\]
where \(\lambda\) is a compensation coefficient. Under this mechanism, the D-model adjusts logits during generation so that, despite extreme steps, the produced tokens tend to align with $P_\text{task}$, potentially yielding a $P_\text{result}$ similar to E-model.

We test this hypothesis via regression, computing the Pearson correlation coefficient \(r\) between \(d_i^{t}\) and \(\Delta P_{\text{token}}^{t}(X = x_i)\). The results (see Figure~\ref{fig:Quota}) show very low correlations—even with \(r<0.1\). Hence, the D-model does not follow this quota compensation to align with \(P_{\text{task}}\). Instead, it may rely on a global plan as we stated in the definition~\ref{definition}, which we leave for future investigation.

\section{Conclusion}
From a novel probabilistic-sampling perspective, we explore the generation process of LLMs by focusing on the alignment between the token-level probability \(P_{\text{token}}\) and the task-level distribution \(P_{\text{task}}\). Through validation on simulated distribution-sampling tasks and systematic evaluation in downstream settings (code generation and recommendation), we identify and characterize two distributional behavior patterns (D-Models and E-Models), observing consistent differences in the diversity–stability trade-off and in distributional alignment. Further analyses,  such as layer-wise probabilities and the quota-compensation hypothesis, provide converging evidence for these two mechanisms. For web scenarios (search, recommendation, and conversational agents), our diagnostics promote transparency of sampling behavior and inform model choice and sampling configuration, enabling a better balance between diversity and reliability under real-world uncertainty.

\section{Acknowledgments}
This work was supported by the Beijing Nova Program under Grants No. 20250484765, the National Natural Science Foundation of China (NSFC) under Grants No. 62276248, the Key Research and Development Program of Xinjiang Uyghur Autonomous Region Grant No. 2024B03026, the Strategic Priority Research Program of the CAS under Grants No.XDB0680302, and the Youth Innovation Promotion Association CAS under Grants No. 2023111.


\bibliographystyle{ACM-Reference-Format}
\balance
\bibliography{sample-base}

\appendix

\section{More Results in Model Validation}
In this section, we present further experimental results from validation experiments in Section~\ref{sec:Validation}.

\subsection{Token Distribution Fluctuations}
\begin{figure*}
  \centering
  \includegraphics[width=\linewidth]{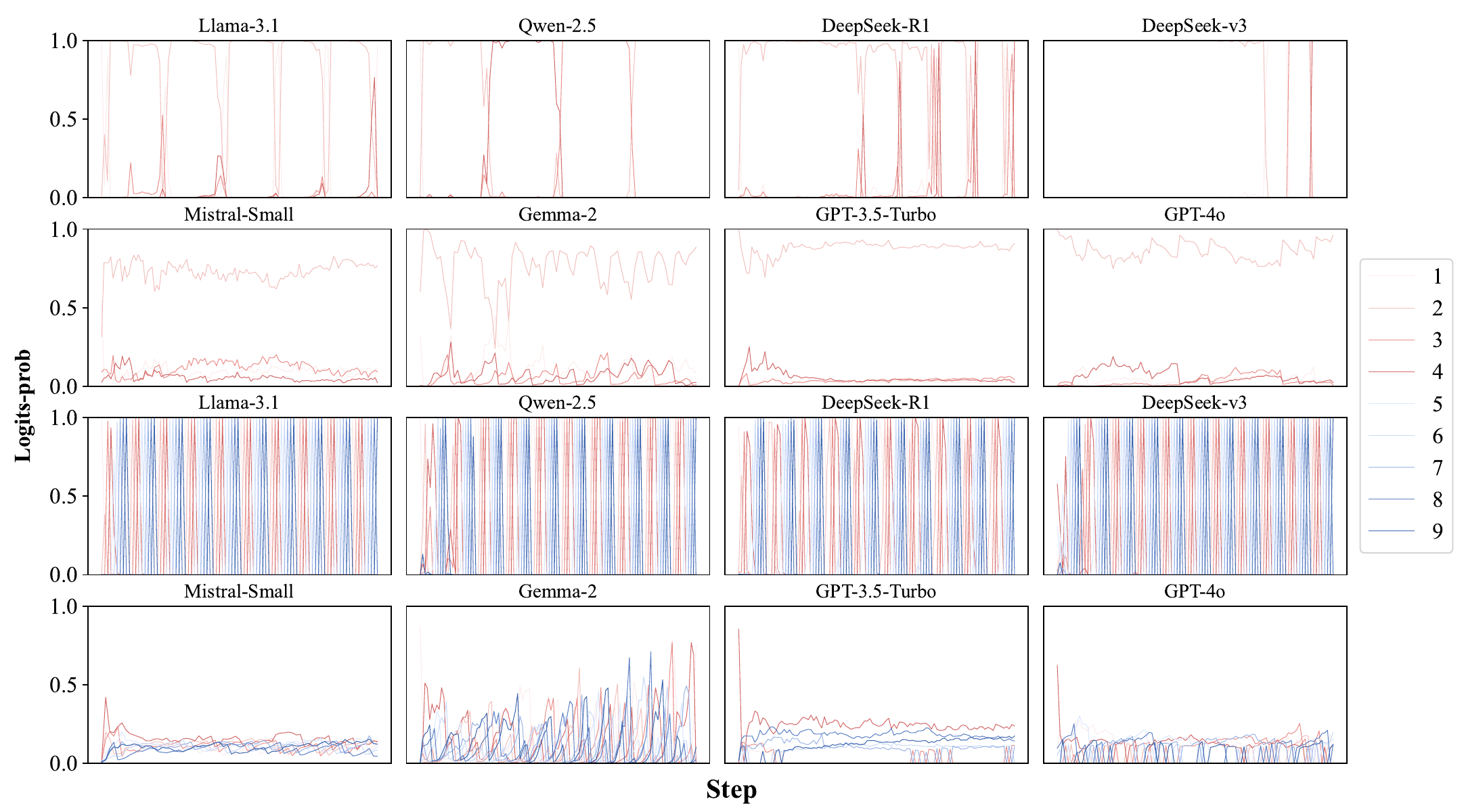}
  \caption{$P_\text{token}$ at each step for each model. The first two rows correspond to extreme distribution task, while the latter two rows pertain to flat distribution task.}
  \Description{}
  \label{fig:app-logits}
\end{figure*}

As shown in the Figure~\ref{fig:app-logits}, the $P_\text{token}$ for each model at every step reveal that the D-model is more volatile, while the E-model's $P_\text{token}$ remains stable and largely aligns with the $P_\text{task}$.

\label{app:logits}

\subsection{Diversity Comparison}

\begin{table}
  \caption{Hamming Distance Comparison of D- and E-Models}
  \label{tab:hamming}
  \centering
  \begin{tabular}{lccc}
    \toprule
    & & \multicolumn{2}{c}{Distribution tasks} \\
    \cmidrule(lr){3-4}
    Group & Model & Extreme & Flat \\
    \midrule
    \multirow{4}{*}{D-model} 
      & Llama-3.1 & 0.32 & 0.36 \\
      & Qwen-2.5 & 0.34 & 0.82 \\
      & DS-R1-Qwen-2.5 & 0.29 & 0.72 \\
      & DeepSeek-v3 & 0.19 & 0.86 \\
    \cmidrule(lr){2-4}
    & AVG & 0.29 & 0.69 \\
    \midrule
    \multirow{4}{*}{E-model} 
      & Mistral-Small & 0.33 & 0.45 \\
      & Gemma-2 & 0.47 & 0.60 \\
      & GPT-3.5-Turbo & 0.20 & 0.79 \\
      & GPT-4o & 0.33 & 0.89 \\
    \cmidrule(lr){2-4}
    & AVG & 0.33 & 0.68 \\
    \bottomrule
  \end{tabular}
\end{table}

We evaluate the diversity of model-generated sequences using the Hamming Distance, which measures the number of differing positions between two equal-length sequences. For sequences of unequal length, the shorter one is zero-padded to match the longer. 

The analysis shows that sequence differences (mean 0.69 for D-models, 0.68 for E-models) are notably higher in flat distribution tasks than in extreme ones (0.29 for D-models, 0.33 for E-models). This stems from the inherent properties of the distributions: the flat distribution has a larger value space (\(9\) vs. \(4\)) and a more uniform target probability (maximum 0.2 vs. 0.7), yielding higher sequence randomness. The difference between D-models and E-models remains minor. Notably, Llama-3.1 consistently stays below 0.36, often producing nearly identical or sequential patterns such as ``1,2,3,4,5,6,7,8,9...’’, indicating limited randomness (see Appendix~\ref{examples}).

\subsection{Practical Task Results}
\label{app:mmlu}
In this section, we present the e-score results for different models on subsets of the MMLU dataset, including high school mathematics, global facts, and all. The DS-R1-Qwen-2.5 model was excluded because its generation always starts with the special token \texttt{<think>}. As shown in Table~\ref{tab:mmlu-escore}, the D-model yields higher e-scores than the E-model across most tasks, confirming the coexistence of both types in practice. Notably, Llama-3.1 shows a lower e-score compared with our simulated experiments, which we hypothesize results from this evaluation considering only the first generated token. Llama-3.1 may produce a flatter \(P_{\text{token}}\) distribution at the initial step, with extreme patterns emerging in subsequent generations.

\begin{table}[ht]
\centering
\caption{Comparison of E-scores Across Models. HSM and GF denote high school mathematics and global facts.}
\label{tab:mmlu-escore}
\begin{tabular}{llccc}
\toprule
Group & Model & HSM & GF & all \\
\midrule
\multirow{5}{*}{D-model} 
& Llama-3.1 & 0.561 & 0.679 & 0.734 \\
& Qwen-2.5   & 0.908 & 0.903 & 0.969 \\
& DS-R1-Qwen-2.5 & -     & -     & -     \\
& DeepSeek-v3         & 0.955 & 0.973 & 0.983 \\
& AVG                   & 0.807 & 0.852 & 0.895 \\
\midrule
\multirow{5}{*}{E-model} 
& Mistral-Small & 0.543 & 0.586 & 0.845 \\
& Gemma-2         & 0.815 & 0.859 & 0.906 \\
& GPT-3.5-Turbo         & 0.668 & 0.683 & 0.869 \\
& GPT-4o                & 0.888 & 0.862 & 0.927 \\
& AVG                   & 0.728 & 0.747 & 0.887 \\
\bottomrule
\end{tabular}
\end{table}

\section{Other Experimental Details}
\subsection{Hardware Setup}
We did not employ any quantization methods in this work, nor did we rely on any specific inference frameworks. All model inference was run directly in a local standard environment, with 8 NVIDIA A100 GPUs and Intel Xeon Platinum 8358 CPUs (2.60 GHz).

\subsection{Detailed Metrics}
Pass@k is a metric for evaluating the programming capabilities of generative models. Given the same problem, the model generates k candidate solutions; if at least one candidate solution passes the preset test cases, the problem is deemed a success. Pass@k measures the probability that the model generates at least one correct solution in multiple attempts, and it is typically used to evaluate the overall performance of code generation models in solving programming tasks. In this paper, we adopt the gap between Pass@5 and Pass@1 to measure the diversity of the code generated by the model.

\section{Prompt Template}
\label{app:prompt}

The prompt template of recommendation is:

\begin{tcolorbox}
\textbf{Recommendation Prompt Template:} \\[4pt]
You are a professional movie recommendation specialist, please complete the task based on the following information:

    user profile: [user profile]
    
    Here are some of the movies he/she has watched and their ratings. You are encouraged to learn his/her movie preference from them:
    [history]
    
    Here's a list of candidate movies that he/she is likely to like: [candidates]

    Please analyze the user's preferences based on his/her movie viewing history and select the top 10 movies that are most likely to be liked from the list given above. 
    
    Be sure to select from the candidate list above and do not generate other movies to ensure that the ids are correct.
    
    Please output only ten movie ids and titles according to the format below without any description.
    
    Output format: "The ten recommended movies are: \\n4: Waiting to Exhale (1995), 8: Tom and Huck (1995), 21: Get Shorty (1995), 24: Powder (1995)..."

\end{tcolorbox}

The prompt template of prior analysis is:
\begin{tcolorbox}
\textbf{Prior Prompt Template:} \\[4pt]
Must generate a hundred numbers in \(\mathcal{X}\) randomly, and give the final list (numbers separated by commas) strictly according to the requirements, without adding any text or process description.
\end{tcolorbox}

\section{Sample Examples}
\label{examples}

\begin{table}
\caption{Sampling Examples of Extreme Distribution Task}
  \centering
  \begin{tabular}{lp{5.6cm}}
    \toprule
    Model & Extreme Distribution Examples  \\
    \midrule
      Llama-3.1 & 1, 2, 2, 2, 2, 2, 2, 2, 2, 2, 2, 2, 2, 2, 2, 2, 2, 2, 2, 2, 2, 2, 2, 2, 2, 2, 2, 2, 2, 3, 3, 4, 1, 1, 1, 1, 1, 2, 2, 2, 2, 2, 2, 2, 2, 2, 2, 2, 2, 2, 2, 2, 2, 2, 2, 2, 2, 2, 2, 3, 3, 3, 1, 1, 2, 2, 2, 2, 2, 2, 2, 2, 2, 2, 2, 2, 2, \dots  \\
      Qwen-2.5 & 1, 2, 2, 2, 2, 2, 2, 2, 2, 2, 2, 2, 2, 2, 2, 2, 2, 2, 2, 2, 2, 2, 2, 2, 2, 2, 2, 2, 2, 2, 2, 2, 2, 2, 2, 2, 2, 2, 2, 2, 2, 2, 2, 2, 2, 2, 2, 2, 2, 2, 2, 2, 2, 2, 2, 2, 2, 2, 2, 2, 2, 2, 2, 2, 2, 2, 2, 2, 2, 2, 2, 2, 2, 2, 2, 2, 2, 2, 
     \dots\\
      DS-R1-Qwen-2.5 & 1, 2, 2, 2, 2, 2, 2, 2, 2, 2, 2, 2, 2, 2, 2, 2, 2, 2, 2, 2, 2, 2, 2, 2, 2, 2, 2, 2, 2, 2, 2, 2, 2, 2, 2, 2, 2, 2, 2, 2, 2, 2, 2, 2, 2, 2, 2, 2, 2, 2, 2, 2, 2, 2, 2, 2, 2, 2, 2, 2, 2, 2, 2, 2, 2, 2, 2, 1, 2, 2, 2, 2, 2, 2, 2, 2, \dots \\
      DeepSeek-v3 & 2, 2, 2, 2, 2, 2, 2, 2, 2, 2, 2, 2, 2, 2, 2, 2, 2, 2, 2, 2, 2, 2, 2, 2, 2, 2, 2, 2, 2, 2, 2, 2, 2, 2, 2, 2, 2, 2, 2, 2, 2, 2, 2, 2, 2, 2, 2, 2, 2, 2, 2, 2, 2, 2, 2, 2, 2, 2, 2, 2, 2, 2, 2, 2, 2, 2, 2, 2, 2, 2, 1, 2, 2, 2, 2, 2, 2, 2, \dots \\
    \midrule
      Mistral-Small & 2, 2, 2, 2, 1, 3, 2, 2, 2, 2, 3, 1, 2, 4, 2, 3, 2, 2, 2, 3, 2, 2, 2, 2, 2, 2, 3, 3, 1, 2, 2, 2, 2, 1, 2, 1, 4, 2, 2, 2, 2, 2, 2, 1, 2, 4, 1, 4, 2, 2, 2, 2, 2, 2, 1, 1, 2, 2, 1, 3, 2, 3, 2, 2, 4, 1, 2, 2, 2, 1, 2, 2, 2, 2, 2, 2, 2, 2, \dots \\
      Gemma-2 & 2, 2, 2, 2, 4, 2, 2, 3, 1, 4, 2, 2, 3, 2, 2, 2, 2, 4, 2, 1, 2, 3, 2, 3, 1, 2, 2, 4, 2, 2, 2, 2, 1, 2, 2, 3, 2, 2, 2, 2, 4, 1, 2, 2, 2, 2, 2, 2, 2, 3, 4, 2, 2, 2, 2, 2, 2, 3, 2, 1, 2, 4, 2, 2, 2, 1, 2, 2, 2, 3, 2, 2, 2, 4, 2, 2, 2, 2, \dots \\
      GPT-3.5-Turbo & 2, 2, 2, 3, 2, 2, 2, 2, 2, 2, 2, 2, 2, 2, 2, 2, 3, 2, 4, 2, 2, 2, 2, 2, 2, 2, 2, 2, 2, 2, 2, 2, 3, 2, 2, 2, 2, 2, 2, 2, 2, 2, 2, 2, 2, 2, 3, 2, 2, 2, 2, 2, 2, 2, 2, 2, 1, 2, 2, 2, 2, 2, 2, 2, 2, 2, 2, 2, 2, 2, 3, 2, 2, 2, 2, 2, 2, 3, \dots \\
      GPT-4o & 2, 2, 2, 2, 2, 2, 2, 2, 2, 2, 2, 2, 2, 2, 2, 2, 2, 2, 2, 2, 2, 2, 2, 2, 2, 2, 2, 2, 2, 1, 3, 2, 2, 2, 2, 2, 2, 2, 2, 2, 2, 2, 2, 2, 4, 2, 1, 2, 2, 2, 2, 2, 2, 2, 2, 2, 2, \dots \\
    \bottomrule
  \end{tabular}
  \label{tab:examples-e}
\end{table}

\begin{table}
\caption{Sampling Examples of Flat Distribution Task}
  \centering
  \begin{tabular}{lp{5.6cm}}
    \toprule
    Model  & Flat Distribution Example \\
    \midrule
      Llama-3.1 &  1, 1, 1, 1, 1, 1, 1, 1, 1, 1, 2, 2, 2, 2, 2, 2, 2, 2, 2, 2, 3, 3, 3, 3, 3, 3, 3, 3, 3, 3, 4, 4, 4, 4, 4, 4, 4, 4, 5, 5, 5, 5, 5, 5, 5, 7, 7, 7, 7, 7, 8, 8, 8, 8, 9, 9, 9, 9, 9, 9, 9 \\
      Qwen-2.5 &  1, 4, 3, 2, 9, 6, 1, 1, 9, 5, 7, 2, 4, 8, 9, 3, 2, 7, 8, 4, 3, 6, 1, 8, 7, 1, 5, 7, 2, 3, 1, 4, 6, 8, 9, 6, 1, 9, 7, 2, 5, 8, 7, 4, 3, 9, 4, 7, 1, 2, 9, 6, 1, 7, 2, 4, 3, 5, 8, 9, 9, 7, 5, 1, 6, 1, 9, 9, 3, 6, 4, 1, 2, 7, 4, 8, 5,  \dots\\
      DS-R1-Qwen-2.5 & 1, 2, 3, 4, 4, 5, 6, 7, 8, 9, 1, 1, 2, 3, 4, 4, 5, 6, 7, 8, 9, 1, 2, 3, 4, 4, 5, 6, 7, 8, 9, 1, 1, 2, 3, 4, 5, 6, 7, 8, 9, 1, 2, 3, 4, 4, 5, 6, 7, 8, 9, 1, 1, 2, 3, 4, 5, 6, \dots\\
      DeepSeek-v3 &  4, 5, 9, 1, 2, 4, 7, 8, 4, 6, 3, 4, 9, 2, 7, 1, 4, 5, 6, 8, 3, 4, 9, 1, 2, 4, 7, 8, 4, 6, 3, 4, 9, 2, 7, 1, 4, 5, 6, 8, 3, 4, 9, 1, 2, 4, 7, 8, 4, 6, 3, 4, 9, 2, 7, 1, 4, 5, 6, 8, 3, 4, 9, 1, 2, 4, 7, 8, 4, 6, 3, 4, 9, 2, 7, 1, 4, 5, \dots\\
    \midrule
      Mistral-Small &  1, 6, 2, 8, 3, 8, 3, 1, 3, 5, 5, 7, 9, 8, 8, 2, 4, 9, 1, 3, 6, 5, 7, 1, 5, 6, 2, 4, 3, 1, 6, 9, 1, 9, 4, 3, 6, 8, 1, 5, 9, 7, 2, 3, 8, 2, 3, 7, 8, 4, 5, 5, 2, 3, 6, 2, 4, 6, 7, 8, 3, 8, 4, 1, 6, 5, 8, 8, 2\\
      Gemma-2 &  1, 4, 2, 5, 9, 3, 7, 5, 1, 4, 8, 1, 6, 9, 2, 8, 7, 1, 4, 6, 3, 5, 9, 2, 4, 8, 7, 5, 3, 1, 9, 6, 8, 4, 7, 1, 3, 2, 6, 5, 9, 8, 2, 7, 1, 5, 4, 9, 4, 8, 6, 3, 5, 2, 7, 1, 3, 9, 5, 8, 4, 6, 2, 1, 9, 7, 3, 2, 8, 6, 5, 4, 3, 7, 1, 9, 8, 5, \dots\\
      GPT-3.5-Turbo &  4, 4, 3, 4, 6, 9, 8, 6, 5, 4, 8, 8, 4, 4, 6, 6, 5, 8, 3, 4, 4, 6, 4, 8, 8, 4, 4, 9, 4, 4, 4, 4, 8, 4, 4, 8, 4, 1, 4, 5, 4, 8, 4, 8, 4, 4, 4, 2, 4, 4, 8, 8, 4, 8, 5, 6, 4, 4, 8, 6, 3, 8, 8, 6, 4, 5, 4, 4, 4, 6, 9, 3, 8, 4, 4, 4, 4, 4, \dots\\
      GPT-4o & 4, 7, 9, 3, 6, 2, 8, 4, 4, 3, 8, 5, 6, 6, 9, 2, 8, 4, 4, 9, 1, 3, 6, 9, 4, 1, 2, 7, 4, 4, 1, 7, 5, 3, 8, 8, 2, 8, 2, 3, 2, 9, 8, 7, 6, 8, 6, 5, 4, 6, 5, 1, 9, 4, 5, 2, 5, 7, 4, 5, 9, 4, 1, 4, 8, 4, 1, 9, 6, 4, 3, 7, 5, 7, 5, 3, \dots\\
    \bottomrule
  \end{tabular}
  \label{tab:examples-f}
\end{table}

This section shows examples of sampling results for different models, as shown in Table~\ref{tab:examples-e} and Table~\ref{tab:examples-f}.

\section{Related Works}

\subsection{Log Probabilities in LLMs}
The log probabilities output by LLMs have been widely utilized in model interpretability research. On one hand, \cite{DBLP:conf/acl/ChengBYOSM024} and \cite{DBLP:journals/npjdm/GuDLY24} respectively validate the effectiveness of using explicit probability outputs for commonsense question answering and medical predictions. However, the latter reveals that most LLMs exhibit an excessively polarized explicit probability distribution, with an overconfidence in predictions that is independent of correctness, suggesting the presence of biases in the model's intrinsic probability distribution.

On the other hand, numerous studies focus on leveraging LLMs' log probabilities to quantify model uncertainty. \cite{DBLP:conf/nips/0001YPWWY0T24} proposes a benchmark for evaluating LLM uncertainty based on logits probabilities. \cite{portillo-wightman-etal-2023-strength} demonstrates that combining log probabilities across multiple prompts yields more accurate confidence estimates, while \cite{DBLP:journals/corr/abs-2406-03441} maps model uncertainty through the stability of explanation distributions generated by the model. Calibration-based concepts can assess LLM uncertainty; \cite{DBLP:journals/tmlr/LinHE22} compares the calibration effects of verbalized probabilities versus probabilities extracted from logits, and \cite{DBLP:journals/tacl/JiangADN21} confirms that model probabilities significantly deviate from actual accuracy rates in traditional QA tasks. \cite{DBLP:journals/corr/abs-2506-18254} innovatively transforms token probabilities into reinforcement learning reward signals, enhancing the reasoning capabilities of large models and expanding the application domain of Reinforcement Learning with Verifiable Rewards (RLVR). Notably, \cite{DBLP:conf/acl/WanJL25} approximates the model's internal beliefs through direct probability questioning, empirically demonstrating the presence of confirmation bias in chain-of-thought reasoning.

Overall, prior work utilizes log probabilities but seldom connects token-level (\(P_{\text{token}}\)) and task-level (\(P_{\text{task}}\)) analyses, often assuming overconfident outputs while neglecting flat distribution behaviors.

\subsection{Probabilistic Reasoning in LLMs}

Current research investigates the probabilistic reasoning abilities of LLMs from various perspectives. In probabilistic reasoning,
\cite{DBLP:conf/acl/OzturklerMWJ23} introduces a two-stage probabilistic reasoning framework based on structured object sets, aggregating query probabilities across phrase groups to improve complex reasoning. 
\cite{DBLP:journals/corr/abs-2507-08182} models reasoning as an explicit probability distribution in latent space, enabling systematic exploration of alternative reasoning paths. 
\cite{DBLP:journals/corr/abs-2402-09614} evaluates LLMs’ capability to perform probabilistic reasoning on text with quantified uncertainty using a Bayesian modeling approach.

However, a significant disconnect exists between LLMs' probabilistic expressiveness and their true sampling capabilities: while \cite{DBLP:journals/corr/abs-2207-05221} verifies that models can quantify uncertainty through confidence levels and that predictive distributions align with real-world frequencies, empirical studies reveal severe limitations. \cite{vo2025bscoredetectingbiaseslarge} identifies biases in LLMs' ability to generate random numbers; \cite{DBLP:conf/coling/GuPSC25}, \cite{hopkins2023can}, and \cite{DBLP:journals/corr/abs-2506-09998} further point out models' difficulty in faithfully sampling target distributions, yet fail to deeply analyze these defects in conjunction with model generation principles. \cite{DBLP:journals/corr/abs-2406-00092} confirms that LLMs not only struggle to generate unbiased samples but may even amplify human cognitive biases.

This "knowledge-action" discrepancy also manifests in complex tasks: \cite{DBLP:conf/naacl/MeisterGH25} discovers deficiencies when LLMs align with group opinion distributions, with the core contradiction lying in models' ability to describe distributions yet inability to sample from them. Moreover, evaluation metrics based on log probabilities may underestimate the performance gap. More fundamentally, \cite{DBLP:journals/corr/abs-2401-16646} detects incoherence in LLMs' probability judgments through probability identity tests. Recent work attempts to bridge this gap: \cite{DBLP:conf/acl/GuptaCGW0DC25} verifies that, given sufficient contextual examples, LLMs roughly update their estimates of coin bias according to Bayesian rules, while \cite{DBLP:conf/iclr/GillmanAF025} proposes a continuous prior learning method from a probability density estimation perspective, attempting to extend models' ability to model complex distributions. Nonetheless, existing research primarily focuses on comparative analysis between LLMs' generated results and tasks, still lacking systematic analysis of whether the intrinsic token-level distribution $P_{\text{token}}$ can adapt to the task-level distribution $P_{\text{task}}$, and failing to clarify the specific impact mechanisms of sampling capability defects in real-world tasks.

\end{document}